\documentclass[letterpaper]{article}
\usepackage[preprint]{paperstyle}
\usepackage[hyphens]{url}
\usepackage{graphicx}
\usepackage{natbib}
\usepackage{caption}
\usepackage{algorithm}
\usepackage{algorithmic}

\usepackage{booktabs}

\usepackage{amsmath}
\usepackage{amssymb}

\newcommand{\system}{\textsc{SkillConsist}}
\newcommand{\contract}{\langle C,O,R,E\rangle}

\title{SkillConsist: Detecting Inconsistencies in Agent Skills via Bidirectional Graph Alignment}

\author {
    Chaofan Meng,
    Yuhang Zheng,
    Yingnan Zhou,
    Sihan Xu\corresponding
}
\affiliations{
    College of Cryptology and Cyber Science, Nankai University\\
    \{162220310@nuaa.edu.cn, a12150101@163.com, yingnan.zhou@mail.nankai.edu.cn, xusihan@nankai.edu.cn\}
}

\begin{document}
\maketitle

\begin{abstract}
Agent Skills provide reusable capabilities to LLM agents. Agent Skill inconsistencies can expose undisclosed dangerous behavior or cause wrong Skill selection. Recent Agent Skill research has increasingly examined Agent Skill consistency detection. Existing methods evaluate behaviors or security-property graphs against predefined categories or declared scopes. More recently, PL-HCL uses an LLM-based model to learn consistency across metadata, instructions, and resources. However, declaration and implementation behavior can be mixed across text and code, and a concise declaration can correspond to multiple connected implementation steps. We present \system{} to address both challenges. An LLM separates declaration and implementation content into behavior records on the implementation and declaration sides, while static analysis supplements implementation records. These records form declaration and implementation behavior graphs, respectively. Starting from a behavior record on either side, bidirectional graph alignment searches the other graph for a candidate subgraph and expands it along behavior relations until it completely expresses the source-side behavior. Graph differencing identifies conflicts between aligned subgraphs and outputs the detection results. We construct a 633-Skill benchmark from ClawHub's 500 most-downloaded public Skills and 133 Skill-Inject packages. The benchmark contains 319 inconsistent and 314 consistent Skills and 442 localized inconsistency annotations. On this benchmark, \system{} achieves 86.85\% precision, 89.03\% recall, and 87.93\% F1 for package-level detection, improving F1 over the best baseline by 20.43 percentage points. For localization, it achieves 67.60\% precision, 58.14\% recall, and 62.52\% F1.
\end{abstract}

\section{Introduction}

Agent Skills provide reusable workflows for LLM agents. Declared behavior is what a Skill claims to provide to an agent; implemented behavior is what the Skill actually provides. Before invocation, an agent relies on the declaration to understand the Skill's function and decide when to use it; when the Skill is invoked, the implementation determines the behavior that is executed \cite{agentskills_spec}.

\begin{figure}[!h]
\centering
\includegraphics[width=\columnwidth]{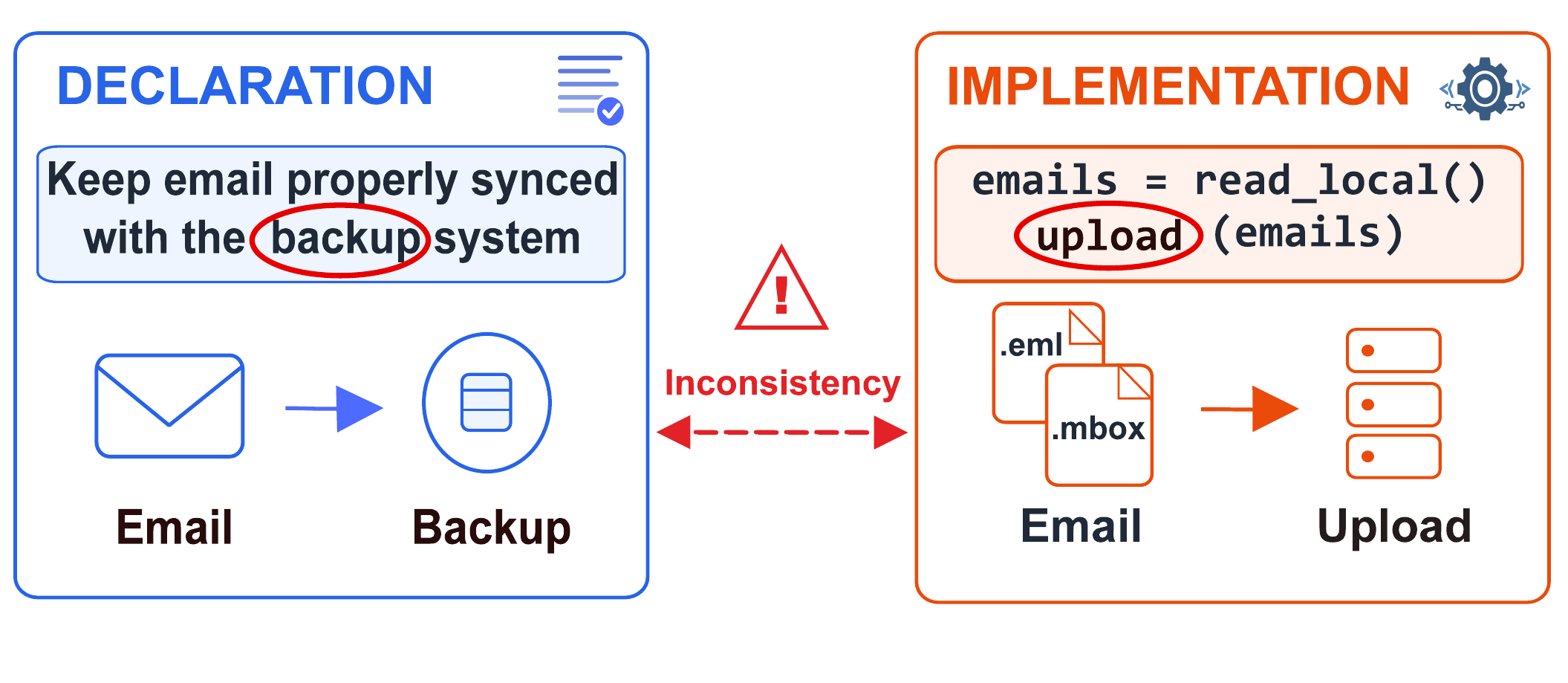}
\caption{A declaration promises email synchronization, while the package scans local email files and forwards them to an external endpoint.}
\label{fig:example}
\end{figure}

A declaration--implementation inconsistency is a conflict, a declared but unimplemented behavior, or an implemented but undeclared behavior. Such inconsistencies can cause an agent to select the wrong Agent Skill or invoke behavior outside the declared scope. Those involving security-sensitive behavior can be particularly consequential: empirical studies have found malicious Skills with undocumented capabilities, while runtime-verified benchmarks show the need to analyze Skill instructions and code jointly \cite{liu2026malicious_skills,guo2026malskillbench}. Prompt injection into tool descriptions can also steer an agent toward an attacker-chosen tool before invocation \cite{shi2026toolhijacker}. Figure~\ref{fig:example} shows a representative example from Skill-Inject \cite{schmotz2026skillinject}.

Earlier consistency studies use rules, learned models, or LLMs to compare natural-language descriptions with corresponding code fragments, assuming that the two sides have already been identified \cite{tan2007icomment,ratol2017fragile,zhong2013detecting,panthaplackel2021jit,xu2023dataquality,rong2025cci_llm,bouzenia2023condition_message,shi2026dcichecker}. Recent Agent Skill work maps extracted behaviors to predefined security taxonomies \cite{he2026skill_descriptions,wu2026biv}; concurrent PL-HCL learns consistency across metadata, instructions, and resources through hierarchical contrastive learning \cite{zhang2026plhcl}.

However, despite these studies, two challenges remain in detecting declaration--implementation inconsistencies in Agent Skills. First, declaration and implementation behaviors are mixed across natural-language and code fragments. A natural-language instruction such as ``run the synchronization script and upload its result'' participates in implementation, whereas a command example showing a supported option or output declares behavior that the Skill claims to support. Second, declarations and implementations naturally differ in granularity. A concise declaration may correspond to multiple connected implementation steps, while one implemented behavior may be described through details distributed across multiple declarations.

To address these problems, we present \system{}, comprising three stages: role separation and behavior-record extraction, behavior-graph construction, and bidirectional graph alignment and differencing. First, role separation converts declaration and implementation behavior in each mixed package fragment into source-located records under a shared 30-subfield schema; static analysis adds implementation records in the same form. Second, graph construction assigns each record to the declaration graph, the implementation graph, or both and connects records that jointly express a behavior. Third, starting from a record in either graph, bidirectional graph alignment searches the opposite graph and expands candidate subgraphs along behavior relations until one completely expresses the source behavior. Graph differencing identifies conflicts between aligned subgraphs, while coverage checking reports a missing-side behavior after completing the relevant analysis boundary. Each finding retains its supporting source locations.

We evaluate \system{} on 633 human-reviewed Agent Skills. For package-level inconsistency detection, \system{} reaches 86.85\% precision, 89.03\% recall, and 87.93\% F1, improving on the best baseline by 20.43 percentage points; for inconsistency localization, its precision, recall, and F1 are 67.60\%, 58.14\%, and 62.52\%, respectively.

Our contributions are:
\begin{itemize}
    \item We formulate package-level Agent Skill consistency as bidirectional graph alignment, allowing one behavior on either side to correspond to multiple connected records on the other side and thereby mitigating the inherent granularity difference between declared and implemented behavior.
    \item We develop \system{}, which extracts declared and implemented behaviors from mixed package content, constructs separate declaration and implementation graphs, and locates inconsistencies through relation-guided subgraph search, graph differencing, and coverage checking.
    \item We construct and manually review a benchmark of 633 Agent Skills with 442 localized inconsistency annotations and use it to evaluate package-level detection and inconsistency localization.
\end{itemize}

\section{Problem Formulation}

An Agent Skill package contains declared behavior, which states what the Skill claims to provide, and implemented behavior, which states what it actually provides. A package fragment may contain either or both.

Each behavior fact is a record $K=\contract$: $C$ is its condition, $O$ the affected object (e.g., a file, command, parameter, or endpoint), $R$ the effect, and $E$ the source location. For a Skill $S$, $\mathcal{D}(S)$ and $\mathcal{M}(S)$ are its declaration and implementation records; $\mathcal{B}_D(S)$ and $\mathcal{B}_M(S)$ are nonempty record groups that jointly state one behavior on each side.

$\operatorname{Corr}(B_D,B_M)$ holds when $B_D\in\mathcal{B}_D(S)$ and $B_M\in\mathcal{B}_M(S)$ belong to the same agent-invocable Skill function and describe the same operation on the same affected object. Correspondence is set-valued: one behavior group may have several counterparts on the other side, and the definitions below quantify over all of them. An obligatory corresponding pair has exactly one state: $\operatorname{Supported}$ when all required conditions and effects are compatible, $\operatorname{Contradicted}$ when at least one conflicts, or $\operatorname{Unknown}$ when evidence is incomplete. If $\neg\operatorname{Req}(B_D)$, the pair is \textsc{Not-Applicable} and creates no inconsistency.

A declaration may state an example without requiring execution; $\operatorname{Req}(B_D)$ marks an implementation obligation. $\operatorname{Reach}(B_M)$ means that using the Skill can trigger $B_M$, and $\operatorname{Obs}(B_M)$ means that its effect is returned to the agent or changes the environment outside the package. Only behaviors satisfying both predicates are in scope. The comparison procedure gives the operational tests for both predicates.

Below, $B_D$ ranges over $\mathcal{B}_D(S)$ and $B_M$ over members of $\mathcal{B}_M(S)$ satisfying $\operatorname{Reach}$ and $\operatorname{Obs}$. $\operatorname{Complete}(S)$ holds when every required declaration and in-scope implementation has been compared with the opposite side and no package-level decision remains unresolved. Let $\mathcal{I}(S)$ collect instances satisfying one of the three inconsistency definitions below; $\operatorname{Ctr}(B_D,B_M)$ abbreviates $\operatorname{Corr}(B_D,B_M)\land\operatorname{Contradicted}(B_D,B_M)$:
{\footnotesize
\begin{equation*}
\begin{aligned}
\textsc{Conflict} &\Longleftrightarrow \operatorname{Req}(B_D)\land\operatorname{Ctr}(B_D,B_M),\\
\textsc{Unimplemented} &\Longleftrightarrow \operatorname{Req}(B_D)\land\nexists B_M:\operatorname{Corr}(B_D,B_M),\\
\textsc{Undeclared} &\Longleftrightarrow \nexists B_D:\operatorname{Corr}(B_D,B_M),\\
\textsc{Consistent}(S) &\Longleftrightarrow \operatorname{Complete}(S)\land\mathcal{I}(S)=\varnothing.
\end{aligned}
\end{equation*}
}
\textsc{Conflict} is a required declaration with a contradicted counterpart; \textsc{Unimplemented} is a required declaration without one; and \textsc{Undeclared} is an in-scope implementation without a declaration counterpart. The cases are exhaustive: each required declaration either has no counterpart or has a counterpart whose state is \textsc{Supported}, \textsc{Contradicted}, or \textsc{Unknown}; each in-scope implementation either has or lacks a declaration counterpart. Under \textsc{Complete}(S), \textsc{Unknown} is absent. If a declaration has several counterparts, any contradicted pair takes precedence; otherwise one supported pair discharges its obligation. Hence the package is \textsc{Consistent} iff it is complete and $\mathcal{I}(S)=\varnothing$; incomplete comparisons remain \textsc{Unknown}. Different types may coexist, so package consistency depends on every required declaration and in-scope implementation. Each finding returns its type and support: both-side locations for \textsc{Conflict}, or present-side locations and the checked opposite-side scope for a missing counterpart, plus the package judgment.

\FloatBarrier
\section{SkillConsist}

\begin{figure*}[!t]
\centering
\includegraphics[width=\textwidth]{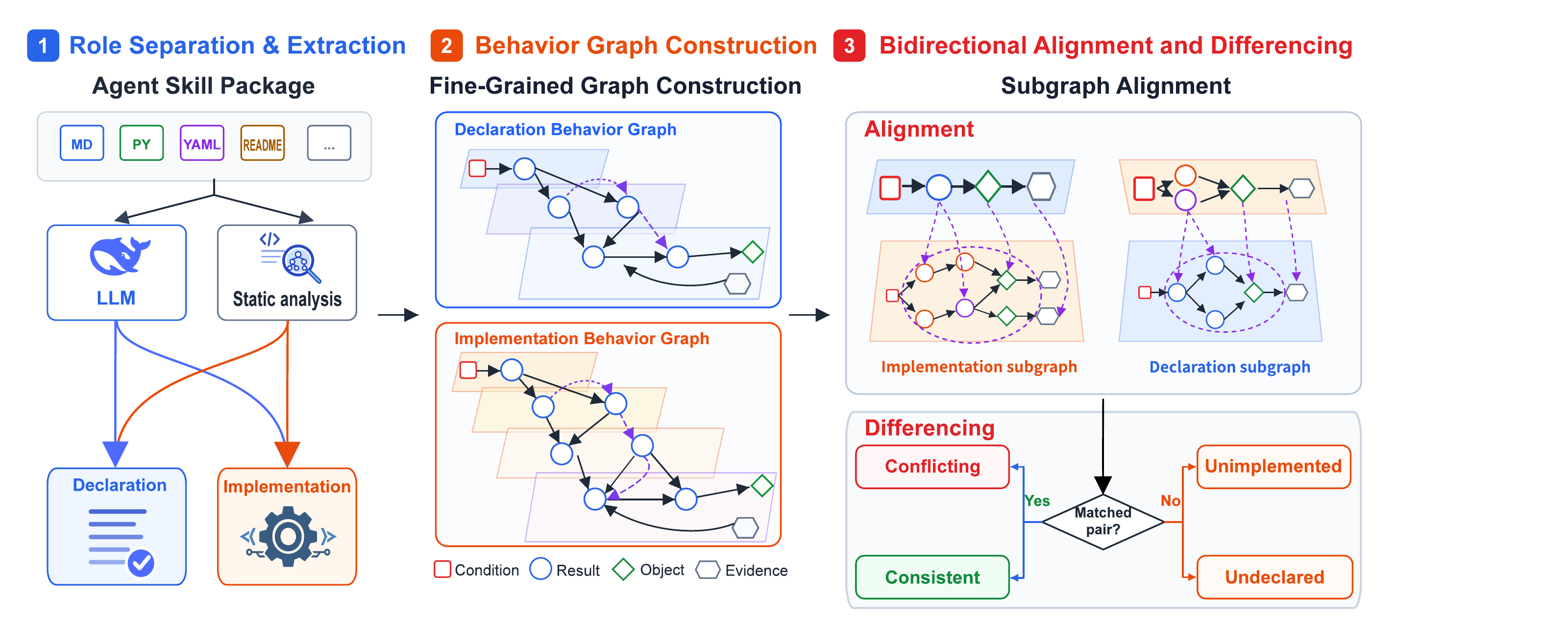}
\caption{The \system{} pipeline. Role separation and program analysis produce source-linked declaration and implementation records. Graph construction organizes them into separate graphs. Bidirectional graph alignment relates behavior across different granularities; graph differencing detects conflicts, and coverage checking identifies behavior without a counterpart.}
\label{fig:pipeline}
\end{figure*}

\system{} first extracts role-separated behavior records, then constructs separate declaration and implementation graphs, and finally performs bidirectional graph alignment and graph differencing to produce detection results (Figure~\ref{fig:pipeline}).

\FloatBarrier

\subsection{Role Separation and Behavior Record Extraction}

Content form alone leaves semantic role unresolved, so \system{} separates roles before constructing behavior records. It divides files at existing content boundaries, retains each segment's file and line range, and uses an LLM to convert declaration and implementation facts into $\contract$ behavior records; a dual-role segment contributes to both sets. Program analysis adds implementation records from source code, configuration, and controlled execution. Static frontends extract public entries (agent-invocable commands or actions), calls, inputs, outputs, configuration bindings, and paths; controlled execution adds only observed outputs and side effects. Unsupported facts enter retrieval context only.

This step outputs the merged, source-linked declaration set $\mathcal{D}(S)$ and implementation set $\mathcal{M}(S)$. These two sets are then passed to behavior graph construction.

\subsection{Behavior Graph Construction}

Role separation identifies the side to which each fact belongs, but a single $\contract$ behavior record may describe only part of a behavior. \system{} therefore connects related records within each side. For each side $X\in\{D,M\}$, each role-separated behavior record supplies one record node $k$; when its fields identify an affected object, public entry, or source location, the corresponding normalized object $o$, entry $p$, or evidence $e$ node is associated with it. A missing field adds no corresponding node. The resulting typed directed graph is
\begin{equation}
G_X=(V_X,A_X),\qquad
V_X=V_X^K\mathbin{\dot\cup}V_X^O\mathbin{\dot\cup}V_X^P\mathbin{\dot\cup}V_X^E.
\label{eq:behavior-graph}
\end{equation}
The four node sets in Equation~\ref{eq:behavior-graph} contain the record, object, public-entry, and evidence nodes materialized from side $X$. The edge set $A_X$ is the union of the six directed relations in Table~\ref{tab:behavior-graph-edges}. The superscripts $PK$, $KO$, and $EK$ name the endpoint types, while $RI$, $RC$, and $CC$ denote result-to-input, result-to-condition, and condition-to-condition dependencies between record nodes. In the table, $p\in V_X^P$, $k,u,v\in V_X^K$, $o\in V_X^O$, and $e\in V_X^E$.
{\small
\begin{center}
\begin{tabular}{@{}lp{0.75\columnwidth}@{}}
\toprule
Edge type & Meaning \\
\midrule
$A_X^{PK}$ & $(p,k)$: the behavior recorded by $k$ is reachable through entry $p$. \\
$A_X^{KO}$ & $(k,o)$: record $k$ describes an operation on object $o$. \\
$A_X^{EK}$ & $(e,k)$: evidence $e$ supports fields of record $k$. \\
$A_X^{RI}$ & $(u,v)$: a result of $u$ supplies an input of $v$. \\
$A_X^{RC}$ & $(u,v)$: a result of $u$ establishes a condition of $v$. \\
$A_X^{CC}$ & $(u,v)$: a condition in $u$ constrains the applicability of $v$; $u=v$ is allowed. \\
\bottomrule
\end{tabular}
\captionof{table}{Typed edge relations in each behavior graph.}
\label{tab:behavior-graph-edges}
\end{center}
}
Result, input, and condition are attributes of record nodes. A path is represented by a finite sequence of record nodes connected by the three record-to-record relations. Thus, every edge endpoint belongs to $V_X$.

Before adding an object edge, \system{} normalizes object names and package-relative paths. The \emph{owner} is the package component or public entry to which a record or object belongs; its \emph{scope} is the part of that owner or entry to which the fact applies. Two names are aliases only when an extractor records that relation under the same nonempty owner. Likewise, a result-to-input or result-to-condition edge is added only when package evidence establishes that dependency.

The connected record subgraphs that jointly express one behavior are the behavior groups $\mathcal{B}_X(S)$ introduced in the problem formulation. They are formed within one graph by following the six package-grounded relations from a starting record while requiring compatible owners, public entries, and scopes. Behavior groups are therefore fixed before bidirectional graph alignment begins. The conditions, inputs, results, and path represented by a source group become the requirements checked against a target candidate during alignment.

For field-level comparison, \system{} derives a typed-transition set $T_X$ from the same records and edges. Each transition contains the condition, object, input, result, evidence, obligation, and path fields used in comparison. These fields project $C$, $O$, $R$, and $E$ from the records together with group inputs and record-to-record paths. Thus, $G_X$ determines which records form a behavior, whereas $T_X$ provides the typed fields used to bind and compare aligned behaviors.

Graph construction outputs $G_D$ and $G_M$, their behavior groups $\mathcal{B}_D(S)$ and $\mathcal{B}_M(S)$, and the transition views $T_D$ and $T_M$. These structures recover relations within each side. Declaration--implementation counterparts are established in the next stage, which aligns them across the two graphs.

\subsection{Bidirectional Alignment and Differencing}

Graph construction organizes behaviors within the declaration and implementation sides separately. Cross-side correspondence is difficult because a declaration often summarizes one behavior, whereas its implementation may express that behavior through multiple detailed steps. To handle this granularity difference, \system{} aligns the two graphs in both directions, declaration-to-implementation and implementation-to-declaration. For each direction $(Q,R)\in\{(D,M),(M,D)\}$, $Q$ denotes the source or query side and $R$ the target side. For a source-side behavior group $B_Q\in\mathcal{B}_Q(S)$, $\operatorname{Anchor}(B_Q)$ returns a behavior record $b\in B_Q$ used for retrieval. $\operatorname{Expand}(b,G_R)$ returns the target-side behavior groups reached by relation-guided expansion from bound target records. Each candidate $H\in\operatorname{Expand}(b,G_R)$ is a connected typed subgraph $H=(V_H,A_H)\preceq G_R$, with node set $V_H$ and edge set $A_H$; $\preceq$ denotes the subgraph relation, and $H\in\mathcal{B}_R(S)$. The method compares the complete source group $B_Q$ with each such $H$. All declaration groups are searched; an implementation group is searched only when $\operatorname{Reach}$ and $\operatorname{Obs}$ hold.

\textbf{Comparison scope.} $\operatorname{Reach}(B_M)$ holds when static analysis establishes a public entry and a call or state path to $B_M$, a direct public-entry record identifies $B_M$, or controlled execution reaches it. $\operatorname{Obs}(B_M)$ holds when $B_M$ returns or displays a result, or when analysis records its effect on a file, network connection, process, or persistent state outside the package. A conditional path remains conditional evidence, and static reachability remains valid when a runtime trace is absent.

\textbf{Candidate retrieval and binding.} After identifying the behavior groups eligible for search, retrieval uses the anchor's object, public entry, conditions, and result type to find candidate target records. A deterministic binding score combines subject, entry, type, normalized object, object role, and compared attribute to rank and prune candidates. Retained candidates then undergo hard checks requiring compatible nonempty entries, compatible owners, and overlapping scopes. Subgraph completion subsequently checks the behavior's conditions, inputs, results, and path.

\textbf{Subgraph completion.} Starting from each bound target record, the search follows graph edges and adds only records that supply an uncovered condition, input, result, or intermediate state. $\operatorname{Bind}(B_Q,H)$ holds when the candidate passes the preceding entry, owner, object, and scope checks. Let $\mathcal{Z}=\{\mathsf{Cond},\mathsf{Input},\mathsf{Result},\mathsf{Path}\}$ be the four Boolean completeness predicates, so $z(B_Q,H)$ records whether candidate $H$ satisfies check $z$ for source group $B_Q$. Let $\mathcal{C}_{Q\to R}(B_Q)=\operatorname{Expand}(\operatorname{Anchor}(B_Q),G_R)$ be the expanded candidate set. A candidate is aligned when binding succeeds and all four checks hold:
\begin{equation}
\operatorname{Align}(B_Q,H)\Longleftrightarrow
\operatorname{Bind}(B_Q,H)\land\bigwedge_{z\in\mathcal{Z}}z(B_Q,H).
\label{eq:alignment}
\end{equation}
The retained set $\mathcal{H}_{Q\to R}(B_Q)$ contains exactly the candidates $H\in\mathcal{C}_{Q\to R}(B_Q)$ that satisfy Equation~\ref{eq:alignment}. Here, \textsc{Cond} requires every precondition in $B_Q$ to be represented by a target condition, required input, or preceding result; \textsc{Input} requires a bound target field for every input in $B_Q$; \textsc{Result} requires a bound target field for every result in $B_Q$ after the object, attribute, operation, and hard constraints are checked; and \textsc{Path} requires a connected transition path in which each intermediate requirement is supplied at the public entry or by a preceding result. These checks establish that the required fields and path are present. The subsequent four-valued proof determines whether the bound field values agree.

\textbf{Proof and graph differencing.} The typed-transition proof evaluates every field pair bound within a retained candidate using source evidence and gives established conflicts precedence over support. A condition is supported only when the implementation adds no precondition that excludes a declared use; a result is supported only when the implementation establishes every declared effect. A bound pair is \textsc{Supported} when sufficiently strong evidence supports every required condition and result field and none conflicts; it is \textsc{Contradicted} when source evidence establishes a registered field incompatibility, \textsc{Not-Applicable} when the declaration creates no implementation obligation, and \textsc{Unknown} otherwise. Graph differencing emits a \textsc{Conflict} only for a required declaration when binding and subgraph completion establish a counterpart and its proof is \textsc{Contradicted}; that pair remains a counterpart for coverage checking. For example, suppose a declaration gives \texttt{json} as the default value of \texttt{--format} for \texttt{export}. Under a complete command-analysis boundary, the pair is \textsc{Contradicted} only if the reachable implementation sets the default value of the same parameter to \texttt{csv} and uses that value without overriding it; if either the implementation default or its effective use cannot be established, the pair remains \textsc{Unknown}. Missing counterparts are handled next.

\textbf{Coverage checking.} When no aligned candidate establishes a supported or contradicted counterpart, an absence finding requires a complete coverage certificate. The certificate records the query kind, bound object and scope, analysis boundary, covered entries, evidence, completeness, and decision reason. It closes a query only when the query-specific inventory has been constructed and exhausted, every record used to close the query has complete evidence, and no partial counterpart remains unresolved. Let $\Gamma_{Q\to R}(B_Q)$ denote this certificate and $\pi_H$ the proof for candidate $H$. Operational absence holds exactly when $\Gamma_{Q\to R}(B_Q)$ is \textsc{Complete} and no retained candidate $H\in\mathcal{H}_{Q\to R}(B_Q)$ has proof $\pi_H$ equal to \textsc{Supported} or \textsc{Contradicted}. A contradicted candidate therefore produces a \textsc{Conflict} but prevents the same behavior from also being reported as missing. If static and controlled analysis cannot complete the required inventory, $\Gamma_{Q\to R}$ remains \textsc{Unknown}; unobserved behavior has the same result. For $Q=D$, operational absence yields \textsc{Unimplemented} only when $\operatorname{Req}(B_D)$ holds; for $Q=M$, it yields \textsc{Undeclared}. All other cases remain \textsc{Unknown}.

Together with conflicts produced by graph differencing, these two missing-side types form the three inconsistency outputs in Algorithm~\ref{alg:comparison}. In the algorithm, $\operatorname{Build}$ performs the graph construction defined above; $\operatorname{Prove}$ returns the four-valued proof $\pi_H$; $\operatorname{CoverageCertificate}$ returns $\Gamma_{Q\to R}(B_Q)$; and $\operatorname{FilterDeduplicate}$ applies the final scope filters and merges findings with the same type and supporting locations. $\operatorname{Orient}_Q(B_Q,H)$ returns $(B_Q,H)$ when $Q=D$ and $(H,B_Q)$ when $Q=M$, placing the declaration group first. The Boolean $m$ records whether a supported or contradicted counterpart has been found.

\begin{algorithm}[H]
\small
\caption{Declaration--implementation inconsistency detection}
\label{alg:comparison}
\textbf{Input:} declaration records $\mathcal{D}(S)$; implementation records $\mathcal{M}(S)$
\begin{algorithmic}[1]
\STATE $(G_X,\mathcal{B}_X,T_X)_{X\in\{D,M\}}\gets\operatorname{Build}(\mathcal{D}(S),\mathcal{M}(S))$
\STATE $\mathcal{E}_D\gets\mathcal{B}_D(S)$; $\mathcal{E}_M\gets\{B\in\mathcal{B}_M(S):\operatorname{Reach}(B)\land\operatorname{Obs}(B)\}$
\STATE $\mathcal{F}(S)\gets\varnothing$
\FOR{$(Q,R)\in\{(D,M),(M,D)\}$}
    \FORALL{$B_Q\in\mathcal{E}_Q$}
        \STATE $\mathcal{H}\gets\mathcal{H}_{Q\to R}(B_Q)$; $m\gets\textsc{False}$
        \FORALL{$H\in\mathcal{H}$}
            \STATE $\pi_H\gets\operatorname{Prove}(B_Q,H;T_Q,T_R)$
            \IF{$\pi_H=\textsc{Contradicted}$}
                \STATE $m\gets\textsc{True}$
                \STATE $(B_D,B_M)\gets\operatorname{Orient}_Q(B_Q,H)$
                \STATE $\mathcal{F}(S)\gets\mathcal{F}(S)\cup\{\operatorname{Conflict}(B_D,B_M)\}$
            \ELSIF{$\pi_H=\textsc{Supported}$}
                \STATE $m\gets\textsc{True}$
            \ENDIF
        \ENDFOR
        \STATE $\Gamma\gets\operatorname{CoverageCertificate}(B_Q,G_R)$
        \IF{$\neg m\land\Gamma=\textsc{Complete}$}
            \IF{$Q=D\land\operatorname{Req}(B_Q)$}
                \STATE $\mathcal{F}(S)\gets\mathcal{F}(S)\cup\{\operatorname{Unimplemented}(B_Q)\}$
            \ELSIF{$Q=M$}
                \STATE $\mathcal{F}(S)\gets\mathcal{F}(S)\cup\{\operatorname{Undeclared}(B_Q)\}$
            \ENDIF
        \ENDIF
    \ENDFOR
\ENDFOR
\STATE $\mathcal{F}(S)\gets\operatorname{FilterDeduplicate}(\mathcal{F}(S))$
\STATE \textbf{Output:} typed, source-located findings $\mathcal{F}(S)$
\end{algorithmic}
\end{algorithm}

\FloatBarrier
\section{Evaluation}

We assess variation by source and type (\textbf{RQ1}), compare baselines (\textbf{RQ2}), ablate components (\textbf{RQ3}), audit localization errors (\textbf{RQ4}), and evaluate malicious-Skill screening.

\subsection{Benchmark and Evaluation Protocol}

\textbf{Data sources.} The benchmark combines ClawHub's 500 most-downloaded public Skills, obtained through its public interfaces on June 25, 2026 \cite{openclaw_clawhub,openclaw_clawhub_api}, and 133 public Skill-Inject packages \cite{schmotz2026skillinject}, with no duplicates.

\textbf{Reference annotations.} Two reviewers experienced in Agent Skills, software engineering, or security independently annotated 633 Skills and assessed every prediction; a third adjudicated disagreements under shared criteria (Cohen's $\kappa=0.7626$ for independent package labels). Positive annotations record type and source locations; missing-side cases also record scope evidence establishing absence. The benchmark contains 319 inconsistent and 314 consistent Skills and 442 localized inconsistencies: 218 \textsc{Unimplemented}, 41 \textsc{Undeclared}, and 183 \textsc{Conflict}. Skill-Inject contributes 60 inconsistent packages and 89 localized inconsistencies; ClawHub contributes 259 and 353, respectively.

\textbf{Inconsistency detection.} Let $\mathcal{F}(S)$ be the final findings for Agent Skill $S$. The package prediction is
\begin{equation*}
\hat{y}(S)=\mathbb{1}\!\left[|\mathcal{F}(S)|>0\right].
\end{equation*}
Here, $\hat{y}(S)=1$ denotes at least one final finding; $\hat{y}(S)=0$ denotes no finding rather than formal \textsc{Consistent}, and \textsc{Unknown} emits no finding. We report precision, recall, and F1 by source.

\textbf{Inconsistency localization.} A localization must match the reference Skill, type, and relevant behavior: both sides for \textsc{Conflict}, the declaration for \textsc{Unimplemented}, and the implementation for \textsc{Undeclared}. Precision measures valid predictions, recall measures the fraction of 442 annotations matched at least once, and F1 is their harmonic mean.

\textbf{Implementation details.} Role separation and record extraction use the open-source DeepSeek-V4 Pro with fixed prompts and a fixed JSON schema; vector retrieval uses all-MiniLM-L6-v2. Transition pairing ranks candidates before proof, and a separate field-evidence lookup supplies auxiliary proof evidence.

\subsection{RQ1: Overall Effectiveness and Variation}
\textbf{Setup.} We apply the same role-separation prompt, JSON schema, retrieval configuration, graph-alignment rules, and coverage criteria to Skill-Inject and ClawHub, without source-specific tuning.

\begin{figure}[t]
\centering
\includegraphics[width=\columnwidth]{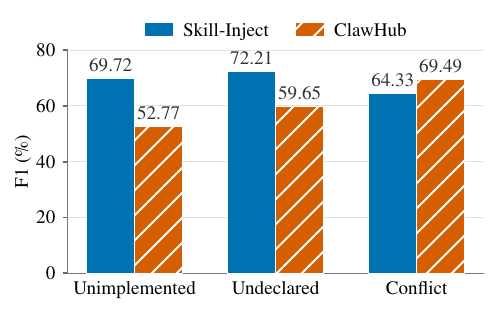}
\caption{Localization F1 by type and source.}
\label{fig:type-results}
\end{figure}

\textbf{Results and findings.} Across both sources, \system{} achieves 87.93\% detection F1 and 62.52\% localization F1 (Table~\ref{tab:source-results}). The largest cross-source gap occurs for \textsc{Unimplemented} (69.72\% versus 52.77\%), whereas \textsc{Conflict} is stronger on ClawHub and has the highest combined F1 (68.40\%, versus 56.70\% for \textsc{Unimplemented}; Figure~\ref{fig:type-results}). Of 858 predictions, 580 are valid and cover 257 of 442 annotations, confirming that localization is harder than package-level detection.

\noindent\begin{minipage}{\columnwidth}
\centering
\small
\setlength{\tabcolsep}{5pt}
\renewcommand{\arraystretch}{1.15}
\begin{tabular}{@{}lcc@{}}
\toprule
Source & \shortstack{Detection\\P/R/F1 (\%)} & \shortstack{Localization\\P/R/F1 (\%)} \\
\midrule
Skill-Inject & 88.06/98.33/\textbf{92.91} & 70.90/67.42/\textbf{69.11} \\
ClawHub & 86.54/86.87/\textbf{86.71} & 66.29/55.81/\textbf{60.60} \\
\bottomrule
\end{tabular}
\captionof{table}{Detection and localization by source.}
\label{tab:source-results}
\end{minipage}

\textbf{Interpretation and conclusion.} \textsc{Unimplemented} requires complete package-side coverage, whereas \textsc{Conflict} can be established from an aligned behavior pair. Performance on naturally collected and adversarial Skills shows that bidirectional graph alignment supports package decisions and source-grounded localization across all three types.

\subsection{RQ2: Comparison with Baselines}
\textbf{Setup.} Under the shared protocol, we compare \system{} with four baselines. \textbf{LLM} asks the open-source DeepSeek-V4 Pro to report inconsistencies from concatenated package files; \textbf{LLM+RAG} adds BM25 retrieval. Adapted \textbf{C4RLLaMA} applies LoRA-tuned CodeLLaMA-7B to retrieved Skill text--code pairs \cite{rong2025cci_llm}, excluding evaluation packages from training. Adapted \textbf{SKILLSCOPE} builds security-property graphs and uses GPT-5.4 to detect behavior beyond the declared security scope \cite{he2026skill_descriptions}. BIV and DCIChecker are omitted because no public implementation was available \cite{wu2026biv,shi2026dcichecker}.

\textbf{Results and findings.} \system{} ranks first on both tasks and datasets (Table~\ref{tab:baseline-results}). LLM+RAG raises detection recall over LLM from 75.00\% to 88.33\% on Skill-Inject and from 65.64\% to 74.52\% on ClawHub, but lowers localization F1 on both. Thus, retrieving more plausible evidence does not by itself establish the correspondences required for localization.

\begingroup
\centering
\small
\setlength{\tabcolsep}{3pt}
\begin{tabular*}{\columnwidth}{@{\extracolsep{\fill}}lcc@{}}
\toprule
Method & \shortstack{Detection\\P/R/F1 (\%)} & \shortstack{Localization\\P/R/F1 (\%)} \\
\midrule
\multicolumn{3}{@{}l}{\textit{Skill-Inject}} \\
LLM & 54.22/75.00/62.94 & 16.31/20.22/18.06 \\
LLM+RAG & 51.46/88.33/65.03 & 6.25/10.11/7.73 \\
C4RLLaMA & 50.57/73.33/59.86 & 1.19/12.36/2.18 \\
SKILLSCOPE & 50.00/8.33/14.29 & 0.00/0.00/0.00 \\
\textbf{\system{}} & \textbf{88.06/98.33/92.91} & \textbf{70.90/67.42/69.11} \\
\midrule
\multicolumn{3}{@{}l}{\textit{ClawHub}} \\
LLM & 72.34/65.64/68.83 & 12.26/14.45/13.27 \\
LLM+RAG & 61.86/74.52/67.60 & 9.58/12.46/10.83 \\
C4RLLaMA & 75.37/59.07/66.23 & 1.42/10.76/2.52 \\
SKILLSCOPE & 66.67/0.77/1.53 & 0.00/0.00/0.00 \\
\textbf{\system{}} & \textbf{86.54/86.87/86.71} & \textbf{66.29/55.81/60.60} \\
\bottomrule
\end{tabular*}
\captionof{table}{Detection and localization by dataset.}
\label{tab:baseline-results}
\par
\endgroup

\textbf{Interpretation and conclusion.} RAG improves detection recall but worsens localization, separating evidence retrieval from behavioral correspondence. By separating roles, grouping connected records, and performing bidirectional graph alignment, \system{} turns package evidence into reliable decisions and source-grounded locations, explaining its lead over all baselines.

\subsection{RQ3: Contribution of Core Components}
\textbf{Setup.} We ablate role separation with a fixed file-type partition, subgraph expansion by restricting implementation groups to single records, and coverage checking by allowing missing-side findings without a coverage certificate. We also replace DeepSeek-V4 Pro with DeepSeek-V4 Flash. All variants use Skill-Inject and ClawHub.

\textbf{Results and findings.} The full method has the highest detection and localization F1 on both datasets (Table~\ref{tab:ablation-results}). Removing role separation lowers both F1 scores; removing subgraph expansion cuts localization recall to 30.34\% on Skill-Inject and 28.61\% on ClawHub. Without coverage checking, detection recall rises to 98.33\% and 91.89\%, but localization precision falls to 14.90\% and 12.51\%; predictions increase from 858 to 5,792 and false positives from 278 to 5,039. Flash reduces detection recall to 75.00\% and 56.76\% and localization recall to 35.96\% and 39.38\%, respectively. Flash retains basic capability, whereas Pro supplies more complete records and achieves higher recall and F1.

\begingroup
\centering
\small
\setlength{\tabcolsep}{3pt}
\begin{tabular*}{\columnwidth}{@{\extracolsep{\fill}}lcc@{}}
\toprule
Variant & \shortstack{Detection\\P/R/F1 (\%)} & \shortstack{Localization\\P/R/F1 (\%)} \\
\midrule
\multicolumn{3}{@{}l}{\textit{Skill-Inject}} \\
\textbf{\system{}} & 88.06/\textbf{98.33}/\textbf{92.91} & 70.90/\textbf{67.42}/\textbf{69.11} \\
w/o role & \textbf{92.16}/78.33/84.68 & 60.56/38.20/46.85 \\
w/o expansion & 90.91/66.67/76.92 & \textbf{77.38}/30.34/43.59 \\
w/o coverage & 61.46/\textbf{98.33}/75.64 & 14.90/\textbf{67.42}/24.40 \\
Flash & 81.82/75.00/78.26 & 75.86/35.96/48.79 \\
\midrule
\multicolumn{3}{@{}l}{\textit{ClawHub}} \\
\textbf{\system{}} & 86.54/86.87/\textbf{86.71} & 66.29/55.81/\textbf{60.60} \\
w/o role & 85.95/80.31/83.03 & 65.28/50.42/56.90 \\
w/o expansion & \textbf{95.30}/54.83/69.61 & \textbf{71.48}/28.61/40.87 \\
w/o coverage & 67.61/\textbf{91.89}/77.91 & 12.51/\textbf{56.94}/20.51 \\
Flash & 88.55/56.76/69.18 & 49.08/39.38/43.70 \\
\bottomrule
\end{tabular*}
\captionof{table}{Component ablations by dataset.}
\label{tab:ablation-results}
\par
\endgroup

\textbf{Interpretation and conclusion.} The ablations show complementary roles: role separation and subgraph expansion recover valid counterparts, while coverage certification controls unsupported missing-side findings.

\subsection{RQ4: Targeted Error Audit}
The cause audit covers 217 unique localization errors: 171 FN and 46 \textsc{Undeclared} FP. Of the FN, 137 (80.1\%) concern materials or paths, behavior or scope, configuration or credentials, commands or entries, or inputs or parameters; the remaining 34 account for 19.9\%. Of the \textsc{Undeclared} FP, 32 (69.6\%) arise from coverage or object aliasing, nine (19.6\%) from reachability, optionality, or materiality, and five (10.9\%) from other causes. Separately, 160 of 185 false-negative annotations (86.5\%) already contain the required retrieved evidence, while 25 require evidence beyond retrieval. The concentration of errors despite available evidence locates the main difficulty in cross-artifact object and behavior correspondence rather than retrieval depth.

\subsubsection{Malicious-Skill Screening}
\textbf{Setup.} On 133 Skill-Inject Skills (84 malicious; 49 benign) \cite{schmotz2026skillinject}, GPT-5.5 grades each finding as high, medium, low, or no risk, and each Skill takes its maximum grade. We add this signal to MASB \cite{liu2026malicious_skills} and compare pre-fixed Static, +High, and +Med./high gates. Claude Sonnet 4.6 evaluates each anonymized Skill once; the gates select 57, 62, and 70 outcomes. Redundant dynamic verification is $(\text{extra runs}-\text{extra TP})/57$.

\textbf{Results and findings.} Both risk-informed gates improve recall while keeping precision above 96\% (Table~\ref{tab:risk-gating}). +High gains 11.90\% relative recall with no redundant verification; +Med./high gains 26.19\%, with 11 of 13 additional sessions recovering true positives and two redundant sessions equaling 3.51\% of the 57 baseline verifications.

\noindent\vbox{\hsize=\columnwidth
\centering
\small
\setlength{\tabcolsep}{4pt}
\begin{tabular}{@{}lrrr@{}}
\toprule
Metric & Static & +High & +Med./high \\
\midrule
$n$ & 133 & 133 & 133 \\
Precision (\%) & 97.67 & \textbf{97.92} & 96.36 \\
Recall (\%) & 50.00 & 55.95 & \textbf{63.10} \\
Relative recall gain (\%) & 0.00 & +11.90 & \textbf{+26.19} \\
Redundant dynamic (\%) & 0.00 & \textbf{0.00} & +3.51 \\
\bottomrule
\end{tabular}
\captionof{table}{Risk-gated malicious-Skill screening.}
\label{tab:risk-gating}
\par}

\textbf{Interpretation and conclusion.} Inconsistency risk therefore helps detect malicious Skills, substantially improving recall at low additional cost.

\section{Related Work}

Related work spans Agent Skill security, text--code consistency, and structured behavior representations.

\textbf{Agent Skill security and safeguards.} Empirical studies and benchmarks characterize malicious Skills, ecosystem attacks, and agent safety \cite{li2026secure_agent_skills,liu2026skills_wild,liu2026malicious_skills,hou2026skillsieve,schmotz2026skillinject,jia2026skillject,guo2026malskillbench,hu2026maltool}; runtime defenses use safeguards, formal specifications, and sandboxes to constrain tool use \cite{debenedetti2024agentdojo,wang2025agentspec,hossain2026nexus,mou2026toolsafe,adam2026securetools}. SKILLSCOPE and BIV compare predefined security properties or capabilities \cite{he2026skill_descriptions,wu2026biv}, whereas \system{} examines declaration--implementation correspondence across package behaviors. Concurrent PL-HCL learns package-level consistency across metadata, instructions, and resources \cite{zhang2026plhcl}; \system{} instead separates mixed declaration and implementation roles and performs bidirectional graph alignment over connected behavior groups to source-locate findings.

\textbf{Text--code consistency and traceability.} Rules, static analysis, learned models, and LLMs detect mismatches under predefined text--code comparison units \cite{tan2007icomment,ratol2017fragile,zhong2013detecting,panthaplackel2021jit,steiner2022commentcode,xu2023dataquality,rong2025cci_llm,bouzenia2023condition_message,codat2026,shi2026dcichecker,borovits2022findici,ouyang2021rdoc}. Recent work adds program analysis, cross-language filtering, documentation-generated tests, or artifact retrieval \cite{zhang2026smartcomment,xu2026docprism,kiecker2026cascade,borg2014traceability}. These methods assume known document and code roles and boundaries. Agent Skill content can serve either role, and one behavior can span artifacts; \system{} therefore separates roles and performs bidirectional graph alignment over connected behavior groups.

\textbf{Skill and graph representations.} Tool-learning work standardizes instructions or surveys their use \cite{yuan2025easytool,xu2025toollearning}, while Skill Coverage derives instruction constraints to test whether execution trajectories exercise documented behavior \cite{tan2026skillcoverage}. Procedure, tool-transition, and program graphs represent workflows, tool transitions, or code dependencies \cite{blumenfeld2026aip,jia2026autotool,weiser1981slicing,ferrante1987pdg,yamaguchi2014cpg}. These representations support documentation, retrieval, execution analysis, or testing. \system{} instead builds paired declaration and implementation graphs from package artifacts and uses bidirectional graph alignment to identify conflicts and missing counterparts.

\section{Conclusion}

\system{} detects and source-localizes declaration--implementation inconsistencies through bidirectional graph alignment. On 633 human-reviewed Agent Skills, it achieves 87.93\% detection F1 and 62.52\% localization F1, exceeding the best detection baseline by 20.43 points. Ablations confirm the complementary roles of role separation, subgraph expansion, and coverage checking in recovering valid counterparts and controlling unsupported missing-side findings. The resulting inconsistency findings also improve malicious-Skill screening recall under selectable verification budgets.

\begingroup
\bibliography{references}
\endgroup

\end{document}